\pdfoutput=1
\documentclass[11pt]{article}
\usepackage{authblk}
\usepackage[]{ACL2023}

\usepackage{times}
\usepackage{latexsym}

\usepackage[T1]{fontenc}
\usepackage[utf8]{inputenc}

\usepackage{microtype}

\usepackage{inconsolata}

\usepackage{multirow}
\usepackage{graphicx}
\usepackage{ctable}
\usepackage{amsmath}
\usepackage{xcolor}
\usepackage{algorithmic,algorithm}
\usepackage{xcolor,colortbl}
\usepackage{xcolor,pifont}
\newcommand*\colourcross[1]{%
  \expandafter\newcommand\csname #1cross\endcsname{\textcolor{#1}{\ding{55}}}%
}
\colourcross{red}
\title{PersonaPKT: Building Personalized Dialogue Agents via Parameter-efficient Knowledge Transfer}

\author[1]{\normalsize\textbf{Xu Han}}
\author[2]{\textbf{Bin Guo}}
\author[2]{\textbf{Yoon Jung}}
\author[2]{\textbf{Benjamin Yao}}
\author[2]{\textbf{Yu Zhang}}
\author[2]{\textbf{Xiaohu Liu}}
\author[2]{\textbf{Chenlei Guo}}
\affil[1]{\normalsize University of Colorado Boulder}
\affil[2]{\normalsize Amazon Alexa }
\affil[ ]{ \normalsize \texttt {xuha2442@colorado.edu, \{guobg, yoonjun, benjamy, yzzhan,}}

\affil[ ]{\normalsize \texttt {derecliu, guochenl\}@amazon.com}}

\date{}

\begin{document}
\maketitle
\begin{abstract}
Personalized dialogue agents (DAs) powered by large pre-trained language models (PLMs) often rely on explicit persona descriptions to maintain personality consistency. However, such descriptions may not always be available or may pose privacy concerns. To tackle this bottleneck, we introduce PersonaPKT, a lightweight transfer learning approach that can build persona-consistent dialogue models without explicit persona descriptions. By representing each persona as a continuous vector, PersonaPKT learns implicit persona-specific features directly from a small number of dialogue samples produced by the same persona, adding less than 0.1\% trainable parameters for each persona on top of the PLM backbone. Empirical results demonstrate that PersonaPKT effectively builds personalized DAs with high storage efficiency, outperforming various baselines in terms of persona consistency while maintaining good response generation quality. In addition, it enhances privacy protection by avoiding explicit persona descriptions. Overall, PersonaPKT is an effective solution for creating personalized DAs that respect user privacy.
\end{abstract}

\maketitle

\section{Introduction}

Recent advances in large-scale pre-trained language models (PLMs) greatly boost the performance of chit-chat dialogue agents (DAs) in generating understandable and fluent responses. However, a PLM-powered DA can potentially suffer from the lack of a consistent personality \cite{zhang-etal-2018-personalizing, li-etal-2016-persona, lian2022incremental} since it is typically trained on dialogues collected from many different personas (i.e., \textit{persona cross-contamination}). To address the issue, many approaches have been proposed to build more persona-consistent models by conditioning on explicit persona descriptions \cite{zhang-etal-2018-personalizing, DBLP:journals/corr/abs-1901-08149}. These descriptions can steer the response generation and are usually presented in the form of several sentences like "\textit{I love the beach.}", "\textit{I am on a diet now.}". However, such explicit persona statements are rarely available in practice: They require hand-crafted feature designs \cite{madotto-etal-2019-personalizing} and are intractable to be directly extracted from real-world conversations or speakers 
\cite{zhang-etal-2018-personalizing, madotto-etal-2019-personalizing, lee2021generating}. Moreover, explicit persona statements may contain sensitive user information, thereby raising privacy concerns.

\begin{figure}
     \centering
         \includegraphics[width=0.4\textwidth, height=0.18\textheight]{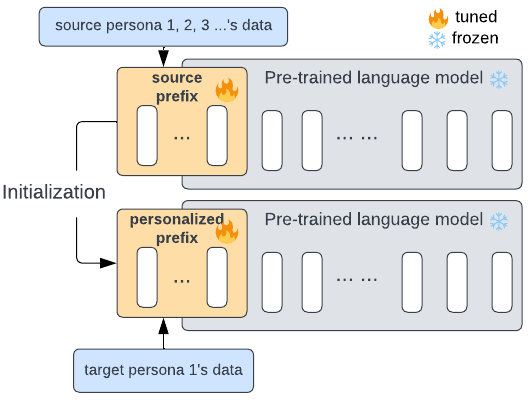}
         \caption{An overview of PersonaPKT.}
         %\Description{}
      \label{fig:PersonaPKT}
      \vspace{-1em}
\end{figure}

In light of this, we introduce PersonaPKT: \textbf {Persona}-based \textbf{P}arameter-efficient \textbf{K}nowledge \textbf{T}ransfer, a lightweight transfer learning approach to build persona-consistent dialogue models without explicit persona descriptions. Inspired by the recent success in lightweight PLM-tuning approaches such as prefix-tuning \cite{li-liang-2021-prefix} and prompt-tuning \cite{lester2021power}, each persona is represented as a continuous vector (i.e., a \textit{personalized prefix}) in PersonaPKT, adding less than 0.1\% trainable parameters compared to a full PLM. PersonaPKT then prepends these personalized prefixes to the PLM with frozen weights to steer the response generation. Instead of utilizing explicit persona descriptions, each personalized prefix learns \textbf{\textit{implicit}} persona-specific features directly from a small number of dialogues produced by the same persona (i.e., low data scenarios). To further improve the response generation quality under such low data scenarios, PersonaPKT first trains one \textit{source prefix} over multiple personas' data agnostically, then uses the \textit{source prefix} to initialize the training of the \textit{personalized prefix} for a target persona (as in Fig~\ref{fig:PersonaPKT}). Through such initialization, PersonaPKT is able to transfer the knowledge learned from various personas to the target \textit{personalized prefix} training, preventing a significant drop in generation quality due to limited personalized training data \cite{madotto-etal-2019-personalizing}. Empirical results show that PersonaPKT is able to build personalized dialogue models with high storage efficiency, outperforming various baselines in terms of persona consistency while maintaining good response generation quality.  In addition, it enhances privacy protection by avoiding explicit persona descriptions.

In practice, PersonaPKT is advantageous in its modularity and extendability. PersonaPKT maintains a separate \textit{personalized prefix} for each persona, making it easily scalable to batch process multiple users while ensuring privacy protection and eliminating the risk of cross-contamination among users' data. Additionally, PersonaPKT's two-stage transfer learning process allows the \textit{source prefix} to be further optimized via different training strategies based on practical needs, showing its great extendability. In this work, we experimented with both data-level (temperature-scaled mixing) and model-level (meta-learning) optimization algorithms to train the \textit{source prefix} to demonstrate the effectiveness of PersonaPKT.   To the best of our knowledge, our work is the first on building personalized dialogue models via parameter-efficient knowledge transfer. As a result, our work makes three unique contributions:

\begin{itemize}
\item {We introduce PersonaPKT, a lightweight transfer learning approach to build personalized dialogue models that respect user privacy.}

\item {We show the great potential of PersonaPKT by further optimizing it via data-level and model-level optimization algorithms.}

\item {We conduct large-scale empirical experiments to show that PersonaPKT builds better personalized dialogue models in terms of both persona consistency and storage efficiency while maintaining good response generation quality.}

\end{itemize}

\section{Related Work}
\subsection{Personalized Dialogue Generation}
Previous studies have shown that when explicit persona descriptions are available, they can be encoded into memory networks (e.g., \citealp{zhang-etal-2018-personalizing}) or appended to dialogue histories to generate persona-consistent responses (e.g., \citealp{DBLP:journals/corr/abs-1901-08149}). However, the utilization of explicit persona descriptions raises privacy concerns as it may require access to personal information. To address this issue, \citet{madotto-etal-2019-personalizing} introduced a persona-agnostic meta-learning (PAML) algorithm that enables the learning of persona-specific dialogue generation models without the need for explicit persona descriptions. Subsequently, several studies have explored this direction using various methodologies \cite{lee-etal-2021-generating, yao2021improving, wu-etal-2021-personalized}. For example, \citet{lee-etal-2021-generating} trained persona-specific models via multi-task meta-learning without any explicit persona descriptions. While the PAML algorithm and its follow-up work demonstrate the feasibility of training persona-consistent dialogue agents without explicit persona description, they still require modifying the entire language model parameters and storing a full copy of the pre-trained model for each persona.  To address this limitation, our approach here provides a more storage-efficient solution for creating effective personalized DAs without the need for explicit persona descriptions.  

\subsection{Parameter-Efficient Knowledge Transfer} 
Instead of modifying all the language model parameters, efficient tuning approaches such as prompt-tuning \cite{lester2021power} and prefix-tuning \cite{li-liang-2021-prefix} optimize a small number of trainable parameters (i.e., prompts/prefixes) for higher flexibility and storage efficiency. However, these parameter-efficient tuning methods occasionally suffer from low training efficiency (e.g., slower convergence rate \cite{su-etal-2022-transferability, xie2022unifiedskg}) and limited model performance compared to full fine-tuning settings \cite{xie2022unifiedskg, li-liang-2021-prefix}. To tackle these issues, many studies have explored the extent to which these parameter-efficient approaches can perform knowledge transfer. \citet{vu-etal-2022-spot} pointed out that prompts/prefixes learned from one or more source tasks were transferrable and could be used to initialize the prompt/prefix for a target task. \citet{su-etal-2022-transferability} further validated such findings and claimed that knowledge transfer can accelerate parameter-efficient tuning methods and improve their performance. In this work, we adapt parameter-efficient knowledge transfer to personalized dialogue modeling, which to our knowledge is the first attempt in this area. 
\section{Methodology}
\subsection{PersonaPKT}
An overview of PersonaPKT is shown in Fig~\ref{fig:PersonaPKT}. Building on the parameter-efficient tuning approach (e.g., prefix-tuning \footnote{In our study, we utilized prefix-tuning. However, PersonaPKT can be adapted to any of the parameter-efficient tuning approaches, such as prompt-tuning. } \cite{li-liang-2021-prefix}) which learns task-specific continuous vectors to condition a frozen pre-trained model, PersonaPKT considers learning different personas as different tasks. Specifically, PersonaPKT adds a trainable persona-specific vector $p_\theta$ for each persona, which we call a \textit{personalized prefix}, depicted by orange blocks in Fig~\ref{fig:PersonaPKT}. The \textit{personalized prefix} is prepended to the task input and the backbone PLM can attend to it as if it were a sequence of "virtual tokens". During training, PersonaPKT only optimizes the personalized prefix while the backbone PLM remains frozen. 

In PersonaPKT, a \textit{personalized prefix} is trained on each persona's dialogue data only. The limited dialogue data per persona will result in a low-data resource scenario, potentially leading to a significant performance drop in terms of dialogue generation quality. In light of this, PersonaPKT is novel in introducing \textit{source prefix} tuning, an extra training stage before \textit{personalized prefix} tuning (Fig~\ref{fig:PersonaPKT}). It first trains one \textit{source prefix} over multiple personas' data agnostically and then uses the \textit{source prefix} to initialize the training of the \textit{personalized prefix} for a target persona. Via such a two-stage transfer learning process, PersonaPKT is able to transfer the knowledge learned from various personas to the target prefix training, preventing the generation quality from dramatically dropping due to limited training data per persona.

\subsection{Optimizing the Training of \textit{Source Prefix}}
PeronsaPKT's two-stage learning process allows the \textit{source prefix} to be further optimized via different training strategies. To validate and fully leverage the benefits of its two-stage learning process, we explored two specific optimization algorithms in terms of data-level (temperature-scaled mixing) and model-level (meta-learning), to train the \textit{source prefix}. 

\subsubsection{Data-level optimization} 
PersonaPKT considers learning the \textit{source prefix} as a multi-task learning process, which involves dialogue data from various personas. As pointed out by previous studies \cite{arivazhagan2019massively}, it is important for multi-task learning to properly mix the data from each task the model should be trained on. We thus utilized temperature-scaled mixing, a common data mixing approach used by multilingual BERT \cite{devlin2018bert} and T5 \cite{raffel2020exploring}, ensuring that the \textit{source prefix} was sufficiently trained on low-resource personas. We implemented temperature scaling with a temperature $T$, where the mixing rate of each persona's data is raised to the power of $\frac{1}{T}$ and then re-normalized so that they sum to 1.

\subsubsection{Model-level optimization}
\citet{madotto-etal-2019-personalizing} proposed a PAML algorithm to learn personalized DA models without any explicit persona descriptions. Inspired by PAML, we employed meta-learning algorithms to further optimize the training of \textit{source prefix}. Specifically, we adapted Reptile \cite{nichol2018reptile}, a widely-used meta-learning algorithm, into our parameter-efficient-tuning-based setting. The adapted algorithm, which we refer to as \textbf{P}arameter-efficient \textbf{P}ersona-agnostic \textbf{Reptile} (PPReptile), is described in Algorithm 1. We further present algorithm details as follows.

We define the persona dataset as $\mathcal{D} = \{P_{1}, P_{2}, ... , P_{z}\}$, where $z$ is the number of personas in  $\mathcal{D}$. 
$\mathcal{D}$ is further split into $\mathcal{D}_{train}$, $\mathcal{D}_{valid}$, $\mathcal{D}_{test}$.  
PPReptile first randomly samples $n$ personas $\rho_{1}, \rho_{2},...,\rho_{n}$ from $\mathcal{D}_{train}$.
With pre-trained parameters $\theta_{pretrain}$ and randomly initialized weights $\theta_{prefix}$, PPReptile then updates the dialogue model $f_{\theta=\theta_{pretrain}\cup\theta_{prefix}}$ with 

\begin{equation}
\label{eq:update}
    \theta_{prefix} \gets \theta_{prefix} + \beta\frac{1}{n}\sum_{i=1}^{n}(W_i - \theta_{prefix})
\end{equation}

, where the gradient $W_{i}$
\begin{equation}
\label{eq:gradient}
    W_{i} = SGD(\mathcal{L}_{\rho_{i}}, \theta_{prefix}, \alpha, k)
\end{equation}

. Here, $\beta$ in Equation~\ref{eq:update}, $\alpha$ in Equation~\ref{eq:gradient} denotes the inner and outer learning rate respectively. $k$ in Equation~\ref{eq:gradient}  denotes $k$ steps of SGD on $\rho_{i}$. $\mathcal{L}_{\rho_{i}}$ is from $L = \{\mathcal{L}_{1}, \mathcal{L}_{1}, ... , \mathcal{L}_{z},\}$, which is the set of loss functions corresponding to all $\rho_{i}$ in $\rho_{1}, \rho_{2},...,\rho_{n}$. Specifically, we use the cross-entropy loss for our response generation task. 

The main difference between PPReptile and vanilla Reptile is that PPReptile updates prefix parameters only, even though the gradient computation still relies on both pre-trained and prefix parameters. 

\begin{algorithm}
\caption{Parameter-efficient Persona-agnostic Reptile}
\begin{algorithmic}[1]
     \REQUIRE model $f_{\theta=\theta_{pretrain}\cup\theta_{prefix}}$ \\
     $\mathcal{D}_{train} = \{\rho_{1}, \rho_{2}, ... , \rho_{z}\}$ \\
     $\alpha$, $\beta$: learning rates \\
     $L = \{\mathcal{L}_{1}, \mathcal{L}_{1}, ... , \mathcal{L}_{z},\}$: set of loss functions corresponding to all potential personas \\
     $k$: inner step number \\
     $n$: persona batch size
    
     \FOR{\textit{iteration} \textbf{in} [1, 2, ...]}
     
        \STATE Sample $n$ personas $\rho_{\{1,2,...,n\}}$ $\sim$ $\mathcal{D}_{train}$ ,
        
        \FOR{\textit{i} \textbf{in} [$1$,$2$,...,$n$]}

            \STATE $W_{i} = SGD(\mathcal{L}_{\rho_{i}}, \theta_{prefix}, \alpha, k)$
            
        \ENDFOR

        \STATE $\theta_{prefix}$ $\gets$ $\theta_{prefix}$ + $\beta\frac{1}{n}\sum_{i=1}^{n}(W_i - \theta_{prefix})$ 
        
    \ENDFOR
\end{algorithmic}
\end{algorithm}

\section{Experiment and Results}
\subsection{The task of personalized dialogue generation}
The task of personalized dialogue generation aims to build dialogue models that are able to generate personalized utterances as response to the input utterance in the context of given dialogue histories. The generated response is expected to not only have good generation quality but also contain information that is consistent with the desired personas. Desired personas are usually provided in the form of several sentences as described in Section 1. In our study, we explore building dialogue models with PersonaPKT for both regular and few-shot personas (i.e., personas with less than 6 dialogues, more details in section 4.2.1).
\subsection{Experiment setup}
\subsubsection{Dataset}

 Our experiments are conducted using PERSONA-CHAT \cite{zhang-etal-2018-personalizing}, a widely-used conversational dataset that contains persona-based dialogues. 
Following \citet{madotto-etal-2019-personalizing}, we first match all dialogues in PERSONA-CHAT by their persona descriptions. 
After examining the distribution of the number of dialogues per persona (Fig~\ref{fig:distribution} in Appendix A.1), we define a few-shot persona if the number of dialogues of that persona is smaller than 6. We had 239 few-shot personas and 1054 regular personas in total. We then randomly set aside 300 regular personas. The remaining 754 regular personas are used as the dataset for the \textit{source prefix} training (\textbf{Part A}).  The 300 regular personas (\textbf{Part B}) and 239 few-shot personas (\textbf{Part C}) are used to train target \textit{personalized prefixes}. There were no overlapped personas among Part A, Part B, and Part C. Within each persona, train, valid and test set are randomly created by dialogue numbers following the ratio of 8:1:1. Table~\ref{tb:dataset} are the basic statistics of our newly-split dataset. 

\begin{table}
\centering
\caption{Statistics of dataset}
\resizebox{0.47\textwidth}{!}
{
\begin{tabular}{lcccc}
\hline
\multicolumn{1}{l|}{}                            & \multicolumn{1}{c|}{\multirow{2}{*}{\# of personas}}                    & \multicolumn{3}{c}{Number of dialogues}                                                                                   \\ \cline{3-5} 
\multicolumn{1}{l|}{}                            & \multicolumn{1}{c|}{}                                                       & \multicolumn{1}{c|}{Train}                   & \multicolumn{1}{c|}{Validation}                   & Test                   \\ \hline
\multicolumn{1}{c|}{Part A$^1$}                    & \multicolumn{1}{c|}{754}                                                    & \multicolumn{1}{c|}{5471}                    & \multicolumn{1}{c|}{774}                          & 774                    \\ \hline
\multicolumn{1}{c|}{Part B$^2$}                    & \multicolumn{1}{c|}{300}                                                    & \multicolumn{1}{c|}{2166}                    & \multicolumn{1}{c|}{304}                          & 304                    \\ \hline
\multicolumn{1}{c|}{Part C$^3$}                    & \multicolumn{1}{c|}{239}                                                    & \multicolumn{1}{c|}{538}                     & \multicolumn{1}{c|}{239}                          & 239                    \\ \hline
\multicolumn{5}{l}{\begin{tabular}[c]{@{}l@{}}$^1$ Part A: for \textit{source prefix} training \\ $^2$ Part B: for \textit{personalized prefixes} training with regular personas \\ $^3$ Part C: for \textit{personalized prefixes} training with few-shot personas\end{tabular}} \\ 
\end{tabular}
}
\label{tb:dataset}
\vspace{-1em}
\end{table}

\subsubsection{Evaluation Metrics}
\noindent \textbf{Automated evaluation} We report F1 score of the generated responses against the human-generated target, which is the standard metric used for PERSONA-CHAT. F1 score can reflect the response quality \cite{madotto-etal-2019-personalizing}. For persona consistency, we report a widely used consistency metric called $C$ score, which was defined by \citet{madotto-etal-2019-personalizing}. They first trained a BERT-based Natural Language Inference (NLI) model to automatically generate NLI annotation between persona descriptions $p_j$ and dialogue utterances $u$ (as Formula (3)). 
\begin{equation}
  NLI (u, p_j) =
    \begin{cases}
      1 & \text{if $u$ entails $p_j$}\\
      0 & \text{if $u$ is independent to $p_j$}\\
      -1 & \text{if $u$ contradicts $p_j$}
    \end{cases}       
\end{equation}

. Based on $NLI(u,p_j)$, the persona consistency score $C$ is then defined as below.
\begin{equation}
C(u) = \sum_{j}^{m}NLI(u,p_j)
\end{equation}

.  The BERT-based NLI model was trained on Dialog NLI \cite{welleck-etal-2019-dialogue} dataset which achieved a test set accuracy of 88.43\%.  In Formula (4), $m$ is the number of sentences in the explicit persona descriptions. $C$ score is shown to be aligned with human-evaluated consistency \cite{madotto-etal-2019-personalizing} and a higher $C$ score means having a more persona-consistent dialogue response. In addition, we report trainable parameter sizes to reflect the storage efficiency of each experiment setting.

~\\

\noindent \textbf{Human evaluation} We also conduct a human evaluation on 377 generated response examples from 50 randomly selected personas in Part B to complement the automatic evaluation results. In accordance with the guidelines provided by \citet{madotto-etal-2019-personalizing}, we requested crowd-sourced workers to assess the fluency (response quality) and persona consistency of the generated response with respect to the dialogue histories and explicit persona descriptions. Specifically, the workers were instructed to rate the response fluency on a 5-point Likert scale ("1 (Not at all)" to "5 (To a great extent)"). For persona consistency, they were required to assign a score of -1, 0, or 1 for \textit{contradicts}, \textit{neutral}, or \textit{consistent}, respectively. 
To ensure the annotation quality, we adopted the following strategies: (1) we only recruited crowd-sourced workers who had an approval rate greater than or equal to 99\% while being located in the United States; (2) two additional annotators further validated the crowd-sourced annotations independently. Following their validation, they resolved any annotation conflicts through discussion or by taking the average score as the final decision.

\subsubsection{Implementation details}
We implemented PersonaPKT using GPT2 \footnote{We only used GPT2 as a testbed to explore the effectiveness of PersonaPKT. PersonaPKT can be adapted to different PLM backbones.} \cite{radford2019language} as the backbone PLM, which has around 345M parameters. A persona-agnostic \textit{source prefix} was first trained with Part A's data. Then we trained \textit{personalized prefixes} for each persona on part B and C. In order to stabilize the training of prefixes, we followed \citet{li-liang-2021-prefix} and employed their parametrization strategy with $k=512$ (the number of persona-specific parameters is less than 0.1\%
of the total GPT2 parameters with $k=512$). When optimizing the training of the \textit{source prefix}, we used $T = 10$ for temperature-scaled mixing. For meta-learning-based optimization, we used learning rates of $\alpha=10^{-4}$, $\beta=3\times10^{-5}$, and batch sizes of $b_{in} = 2$, $b_{out} = 4$ for the inner, outer loops, respectively. When training target \textit{personalized prefixes}, we tuned batch size and learning rate with early stopping on each persona's validation set. During model training, we used the AdamW optimizer \cite{loshchilov2018decoupled} and a linear learning rate scheduler for all the models. We also used beam search with a beam size of 5 when decoding.

\begin{table*}[ht]
\renewcommand{\arraystretch}{1.2}
\caption{Results of automatic evaluation. Significantly underperforming settings are highlighted with \redcross.}
\resizebox{1.06\textwidth}{!}{\begin{tabular}{l|l|lllllllll}
\hline
                                                                                                  &                           & \multicolumn{9}{c}{Automatic Metrics}                                                                                                                                                                                                                                                                                                 \\ \cline{3-11} 
                                                                                                  &                           & \multicolumn{2}{l|}{1-gram F1 ↑}                            & \multicolumn{2}{l|}{2-gram F1 ↑}                            & \multicolumn{2}{l|}{LCS F1 ↑ $^\textbf{1}$}                               & \multicolumn{2}{l|}{C Score ↑}                               & \multirow{2}{*}{\begin{tabular}[c]{@{}l@{}}Trainable \\ Parameter Size ↓ $^\textbf{2}$\end{tabular}} \\ \cline{3-10}
                                                                                                  &                           & \multicolumn{1}{l|}{Part B $^\textbf{3}$} & \multicolumn{1}{l|}{Part C $^\textbf{4}$} & \multicolumn{1}{l|}{Part B} & \multicolumn{1}{l|}{Part C} & \multicolumn{1}{l|}{Part B} & \multicolumn{1}{l|}{Part C} & \multicolumn{1}{l|}{Part B}  & \multicolumn{1}{l|}{Part C} &                                                                                      \\ \hline \rowcolor{lightgray}
\begin{tabular}[c]{@{}l@{}}With \\ descriptions\end{tabular}                     & Persona + Fine-tuning $^\textbf{5}$     & \multicolumn{1}{l|}{\textbf{18.64}}  & \multicolumn{1}{l|}{\textbf{17.42}}  & \multicolumn{1}{l|}{\textbf{6.76}}   & \multicolumn{1}{l|}{5.68}   & \multicolumn{1}{l|}{\textbf{17.69}}  & \multicolumn{1}{l|}{\textbf{16.50}}  & \multicolumn{1}{l|}{0.20}    & \multicolumn{1}{l|}{0.23}   & 1 * 100\%                                                                            \\ \hline
\multirow{7}{*}{\begin{tabular}[c]{@{}l@{}}Without \\descriptions\end{tabular}} & Fine-tuning               & \multicolumn{1}{l|}{18.16}  & \multicolumn{1}{l|}{17.18}  & \multicolumn{1}{l|}{6.27}   & \multicolumn{1}{l|}{\textbf{5.86}}   & \multicolumn{1}{l|}{17.30}  & \multicolumn{1}{l|}{16.22}  & \multicolumn{1}{l|}{\underline{-0.0082} \redcross} & \multicolumn{1}{l|}{\underline{0.0076} \redcross} & 1 * 100\%                                                                            \\
                                                                                                  & Persona\_id + Fine-tuning & \multicolumn{1}{l|}{17.87}  & \multicolumn{1}{l|}{16.84}  & \multicolumn{1}{l|}{5.98}   & \multicolumn{1}{l|}{5.45}   & \multicolumn{1}{l|}{16.89}  & \multicolumn{1}{l|}{15.88}  & \multicolumn{1}{l|}{\underline{0.00} \redcross}    & \multicolumn{1}{l|}{\underline{0.00} \redcross}   & 1 * 100\%                                                                            \\
                                                                                                  & Reptile + Fine-tuning     & \multicolumn{1}{l|}{16.30}  & \multicolumn{1}{l|}{15.40}  & \multicolumn{1}{l|}{4.44}   & \multicolumn{1}{l|}{3.78}   & \multicolumn{1}{l|}{15.20}  & \multicolumn{1}{l|}{14.22}  & \multicolumn{1}{l|}{0.27}    & \multicolumn{1}{l|}{0.21}   & \underline{(N+1) * 100\%} \redcross                                                                        \\
                                                                                                  & Rand init + Prefix-tuning & \multicolumn{1}{l|}{\underline{9.36} \redcross}   & \multicolumn{1}{l|}{\underline{8.12} \redcross}   & \multicolumn{1}{l|}{\underline{1.34} \redcross}   & \multicolumn{1}{l|}{\underline{1.21} \redcross}   & \multicolumn{1}{l|}{\underline{6.74} \redcross}   & \multicolumn{1}{l|}{\underline{4.98} \redcross}   & \multicolumn{1}{l|}{\textbf{0.28}}    & \multicolumn{1}{l|}{\textbf{0.26}}   & N * 0.1\%                                                                            \\
                                                                                                  & PersonaPKT (base)         & \multicolumn{1}{l|}{16.35}  & \multicolumn{1}{l|}{14.91}  & \multicolumn{1}{l|}{4.77}   & \multicolumn{1}{l|}{3.96}   & \multicolumn{1}{l|}{15.39}  & \multicolumn{1}{l|}{13.89}  & \multicolumn{1}{l|}{0.14}    & \multicolumn{1}{l|}{0.11}   & (N+1) * 0.1\%                                                                        \\
                                                                                                  & PersonaPKT (temperature)  & \multicolumn{1}{l|}{16.40}  & \multicolumn{1}{l|}{16.38}  & \multicolumn{1}{l|}{4.53}   & \multicolumn{1}{l|}{4.66}   & \multicolumn{1}{l|}{15.41}  & \multicolumn{1}{l|}{15.15}  & \multicolumn{1}{l|}{0.15}    & \multicolumn{1}{l|}{0.12}   & (N+1) * 0.1\%                                                                        \\
                                                                                                  & PersonaPKT (PPReptile)    & \multicolumn{1}{l|}{16.25}  & \multicolumn{1}{l|}{15.30}  & \multicolumn{1}{l|}{4.69}   & \multicolumn{1}{l|}{3.64}   & \multicolumn{1}{l|}{15.17}  & \multicolumn{1}{l|}{14.01}  & \multicolumn{1}{l|}{0.21}    & \multicolumn{1}{l|}{0.16}   & (N+1) * 0.1\%                                                                        \\ \hline
\multicolumn{10}{l}{$^\textbf{1}$ LCS F1 denotes the longest common subsequence F1;}\\ 
\multicolumn{10}{l}{$^\textbf{2}$ N in this column represents the number of personas;}    \\
\multicolumn{10}{l}{$^\textbf{3}$$^\textbf{4}$ Part B for regular personas while part C for few-shot personas; }\\
\multicolumn{10}{l}{$^\textbf{5}$ The model is trained with explicit persona descriptions while other models not.}\\ 

\end{tabular}}
\label{tb:results_automatic}
\vspace{-1em}
\end{table*}

\subsection{Experiment Settings}
We compare the following 8 training settings: 
\begin{itemize}
\item \textbf{Persona + Fine-tuning}: A GPT2 model fine-tuned on Part A and evaluated on the test set of Part B and C. Explicit persona descriptions were appended to the input utterance during both training and inference. Although this setting utilized explicit persona descriptions, which is different from the underlying assumptions of PersonaPKT (i.e., without explicit persona descriptions), we still include this setting as a point of reference.

\item \textbf{Fine-tuning}: Without any explicit persona descriptions as input, this setting fine-tuned a GPT2 model on Part A in a persona-agnostic manner and evaluated on the test set of Part B and C; 

\item \textbf{Persona\_id + Fine-tuning}: A GPT2 model fine-tuned on Part A and evaluated on the test set of Part B and C. 
Same as Persona + Fine-tuning while only the persona id was appended to the input utterance instead of the explicit persona descriptions;

\item \textbf{Reptile + Fine-tuning}: Multiple persona-specific GPT2 models fine-tuned for each persona respectively in Part B and C. Each persona-specific GPT2 model was trained on dialogues produced by that persona only. Each personalized GPT2 model was initialized using a GPT2 model, which was fine-tuned in a persona-agnostic manner on Part A with the Reptile algorithm; 

\item \textbf{Rand init + Prefix-tuning}: \textit{Personalized prefixes} were trained for each persona respectively in Part B and C. Each \textit{personalized prefix} was trained on dialogues produced by that persona only. Prefix weights were randomly initialized;

\item \textbf{PersonaPKT (base)}: \textit{Personalized prefixes} were trained for each persona respectively in Part B and C. Each \textit{personalized prefix} was trained on dialogues produced by that persona only. Each personalized prefix was initialized using a \textit{source prefix}, which was trained in a persona-agnostic manner on Part A;

\item \textbf{PersonaPKT (temperature)}: \textit{Personalized prefixes} were trained for each persona respectively in Part B and C. Each \textit{personalized prefix} was trained on dialogues produced by that persona only. Each personalized prefix was initialized using a \textit{source prefix}, which was trained in a persona-agnostic manner on Part A's data after temperature-scaled mixing;

\item \textbf{PersonaPKT (PPReptile)}: \textit{Personalized prefixes} were trained for each persona respectively in Part B and C. Each \textit{personalized prefix} was trained on dialogues produced by that persona only. Each personalized prefix was initialized using a \textit{source prefix}, which was trained in a persona-agnostic manner with PPReptile in Algorithm 1 using Part A's data.
\end{itemize}

\begin{table}[ht]
\centering
\renewcommand{\arraystretch}{1.1}
\caption{Results of human evaluation. Significantly underperforming settings are highlighted with \redcross.}
\resizebox{0.55\textwidth}{!}{\begin{tabular}{l|l|ll}
\hline
                                                                                                  &                           & \multicolumn{2}{c}{Human Metrics}                                                  \\ \cline{3-4} 
                                                                                                  &                           & \multicolumn{1}{l|}{Fluency} & \begin{tabular}[c]{@{}l@{}}Persona  \\ Consistency\end{tabular} \\ \hline \rowcolor{lightgray}
\begin{tabular}[c]{@{}l@{}}With \\ descriptions\end{tabular}                     & Persona + Fine-tuning     & \multicolumn{1}{l|}{3.59}    & 0.18                                                           \\ \hline
\multirow{7}{*}{\begin{tabular}[c]{@{}l@{}}Without \\ descriptions\end{tabular}} & Fine-tuning               & \multicolumn{1}{l|}{3.68}    & \underline{0.00080} \redcross                                                        \\
                                                                                                  & Persona\_id + Fine-tuning & \multicolumn{1}{l|}{3.69}    & \underline{-0.024} \redcross                                                         \\
                                                                                                  & Reptile + Fine-tuning     & \multicolumn{1}{l|}{3.62}    & \textbf{0.22}                                                           \\
                                                                                                  & Rand init + Prefix-tuning & \multicolumn{1}{l|}{\underline{2.49} \redcross}    & 0.12                                                           \\
                                                                                                  & PersonaPKT (base)         & \multicolumn{1}{l|}{\textbf{3.71}}    & 0.17                                                           \\
                                                                                                  & PersonaPKT (temperature)  & \multicolumn{1}{l|}{3.46}    & 0.20                                                           \\
                                                                                                  & PersonaPKT (PPReptile)    & \multicolumn{1}{l|}{3.42}    & 0.20                                                           \\ \hline
\end{tabular}}
\label{tb:results_human}
\end{table}

\subsection{Results}
Automatic and human evaluation results are presented in Table~\ref{tb:results_automatic} and Table~\ref{tb:results_human}, respectively. Overall, PersonaPKT outperforms various baselines in terms of persona consistency in both automatic and human evaluation metrics. When explicit persona descriptions are not available, PersonaPKT is capable of achieving significantly higher persona consistency compared to fine-tuning baselines in both automatic and human evaluation metrics, regardless of the optimization strategies (\textbf{PersonaPKT} vs. \textbf{Fine-tuning}, \textbf{Persona\_id+Fine-tuning}). Moreover, \textbf{PersonaPKT (PPReptile)} achieves even higher human-annotated persona consistency than the fine-tuning baseline in scenarios where explicit persona descriptions are available (\textbf{PersonaPKT} vs. \textbf{Persona+Fine-tuning}). 

For response quality, the comparison between \textbf{PersonaPKT} and \textbf{Rand init + Prefix-tuning} further shows the effectiveness of using a \textit{source prefix} to maintain good response generation quality. While PersonaPKT has lower F1 scores than other fine-tuning baselines, \textbf{PersonaPKT (base)} has achieved the highest human-annotated fluency compared to other baselines. This finding aligns with previous studies that indicate F1 measures are not highly correlated with human judgments \cite{madotto-etal-2019-personalizing, liu-etal-2016-evaluate}. 
For completeness, we show generated response examples from PersonaPKT and baseline models in Appendix A.2. 

In addition, PersonaPKT finds a good trade-off between storage efficiency and model performance. Although PersonaPKT performs slightly worse than the \textbf{Reptile + Fine-tuning} baseline in terms of both automatic and human-annotated persona consistency, its storage efficiency is considerably higher, resulting in more practical utility. As shown in Table~\ref{tb:results_automatic},  all our findings in terms of the automatic metrics can be generalized to few-shot personas as well (Part C). 

In conclusion, although PersonaPKT doesn't achieve the highest score in any of the evaluation metrics across the board except human-annotated fluency (Table~\ref{tb:results_automatic}, Table~\ref{tb:results_human}), it overcomes the limitations of all the baseline models, avoiding significantly poor performances in neither response quality, persona consistency nor storage-efficiency (significantly underperforming metrics under each setting were highlighted with \redcross). 

\subsection{Ablation Study}
\begin{figure}[ht]
     \centering
         \includegraphics[width=0.45\textwidth, height=0.2\textheight]{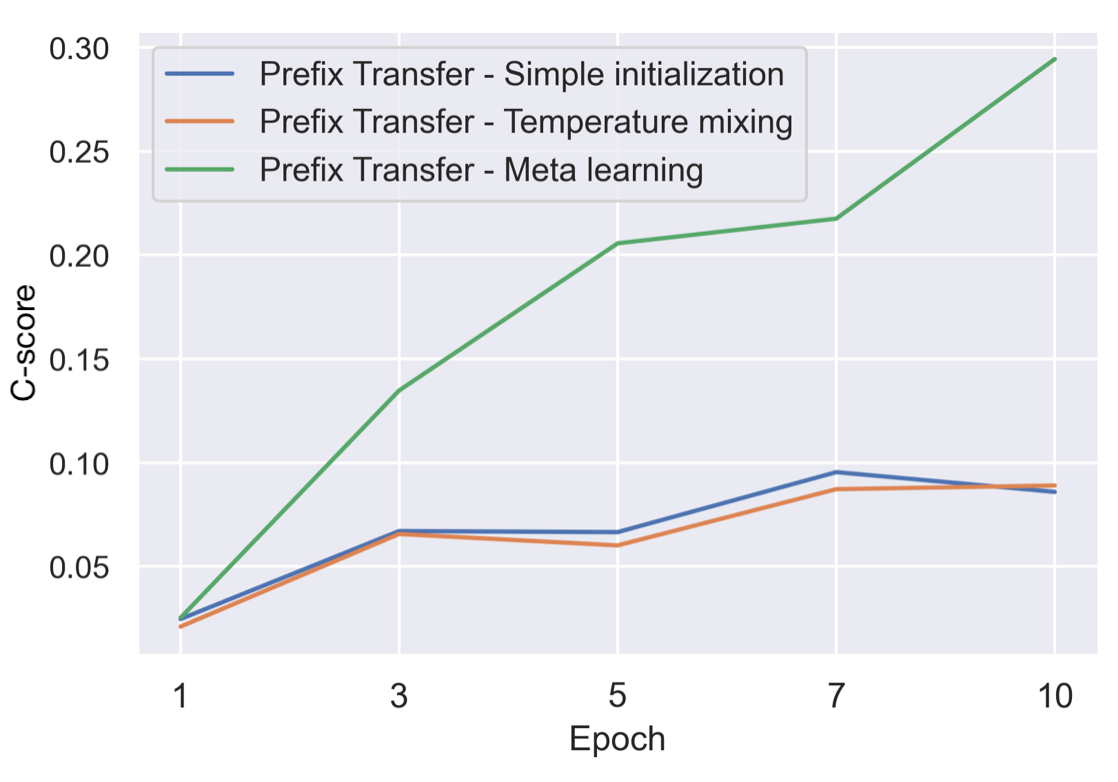}
         \caption{Ablation study results:  how fast the \textit{source prefix} can be adapted to a certain persona with different PersonaPKT optimization strategies.}
         %\Description{}
      \label{fig:ablation}
\end{figure}

In the ablation study, we further evaluate the impact of different PersonaPKT optimization approaches. Specifically, we study how fast the \textit{source prefix} can be adapted to a certain persona when training \textit{personalized prefix}. As shown in Fig~\ref{fig:ablation}, we compared \textbf{PersonaPKT (base)}, \textbf{PersonaPKT (temperature)} and \textbf{PersonaPKT (PPReptile)} in terms of their $C$ score with controlled numbers of epochs. The experiment was conducted on 100 randomly-selected personas from Part B.

As shown in Fig~\ref{fig:ablation}, \textbf{PersonaPKT (PPReptile)} has the best performance: it can achieve the highest $C$ score with the fewest training epochs, demonstrating the effectiveness of utilizing meta-learning to train the \textit{source prefix}. In contrast, Table~\ref{tb:results_automatic} and Table~\ref{tb:results_human} reveal that \textbf{PersonaPKT (base)} and \textbf{PersonaPKT (temperature)} have better response quality than \textbf{PersonaPKT (PPReptile)}. These observations highlight the extendability of PersonaPKT's two-stage transfer training process, which enables the source prefix to be further optimized via various training strategies. Such extendability is valuable in practice as it indicates that engineers can choose or even propose customized optimization strategies to train the source prefix based on their specific needs. For example, if their products prioritize persona consistency over F1 score, PPReptile could be a suitable choice for optimization.

\section{Limitations}
\subsection{Automatic metrics vs. Human annotation} 
\noindent \textbf{ F1 vs. fluency.} As discussed in section 4.4, we have observed that the F1 score cannot be well-correlated with human annotations across almost all experiment settings. This is mostly due to the inflexibility of computing the F1 score sorely based on the similarity between the generated responses and golden references. Such observations echo findings from previous studies \cite{madotto-etal-2019-personalizing, liu-etal-2016-evaluate, deutsch2022re}. In this work, we still utilize F1 score since it’s a standardized metric for PERSONA-CHAT evaluation. However, future work should aim to find a better evaluation metric that can more comprehensively reflect the quality of the generated responses. \\

\noindent \textbf{$C$ score vs. human-annotated consistency.} We observed a significant discrepancy between the $C$ score and human-annotated consistency under the setting of Rand init + Prefix-tuning. Upon further analysis, we discovered that many responses generated from this setting contain repeated sentences with persona-consistent keywords (e.g., as shown in Table~\ref{tb:examples}, Rand init + Prefix-tuning repeated \textit{"he is a preacher"} multiple times). While human annotators tend to ignore these repeated sentences, the $C$ score considers them as highly consistent. Our observations indicate that although the $C$ score has been shown to be a good indicator of persona consistency \cite{madotto-etal-2019-personalizing}, it is still limited under certain experiment settings like Rand init + Prefix-tuning. Similar to the F1 score, a better evaluation metric is needed in the future to more comprehensively reflect the persona consistency of generated responses.

\subsection{Hyperparameter Selection} 
One limitation of PersonaPKT is its sensitivity to hyperparameters. Since PersonaPKT maintains persona-specific prefixes, this modularity of PersonaPKT allows flexibility in optimizing hyperparameters for each persona, such as learning rate, batch size, etc. Despite the advantage, our experiments show that the performance of PersonaPKT is more sensitive to hyperparameters compared to baseline models trained over multiple personas' data agnostically (Persona + Fine-tuning, Fine-tuning, Persona\_id + Fine-tuning). At the same time, persona-specific baseline models trained on a small number of dialogues produced by a single persona (Reptile + Fine-tuning, Rand init + Prefix-tuning) demonstrate similarly high sensitivity to hyperparameters as PersonaPKT. Such observations indicate that this high sensitivity is a common issue in persona-specific models like PersonaPKT. Strategies for improving their robustness to hyperparameters can be a potential study area in the future.

\section{Discussion}

In practice, PersonaPKT is advantageous in its modularity and extendability.  Due to its high storage efficiency, PersonaPKT is advantageous when there are a large number of user-specific models that need to be maintained independently. PersonaPKT offers not only scalability to batch process multiple users, but also enhances user privacy by avoiding cross-contamination between different users' data. Moreover, PersonaPKT’s 
great extendability allows engineers to adopt various source prefix optimization strategies, parameter-efficient tuning approaches or even PLM backbones based on practical needs. Lastly, it enhances privacy protection by avoiding the use of explicit persona descriptions. All of these advantages of PersonaPKT make it a valuable contribution to personalized DA training in the industry.

\section{Conclusion}
We present PersonaPKT, a lightweight transfer learning approach for building persona-consistent dialogue models without the need for explicit persona descriptions. By representing each persona as a continuous vector, PersonaPKT learns implicit persona-specific features directly from dialogue samples produced by the same persona, with less than 0.1\% trainable parameters added for each persona on top of the PLM backbone. Its two-stage learning process provides training flexibility, allowing for various optimization strategies to further enhance the training of the source prefix. PersonaPKT offers potential in terms of privacy protection and batch processing of multiple users. Future work will explore different optimization strategies and generalize PersonaPKT to additional applications.

% Entries for the entire Anthology, followed by custom entries
\bibliography{anthology,custom}
\bibliographystyle{acl_natbib}

\appendix
\section{Appendix}

\begin{figure*}
     \centering
         \includegraphics[width=0.5\textwidth, height=0.26\textheight]{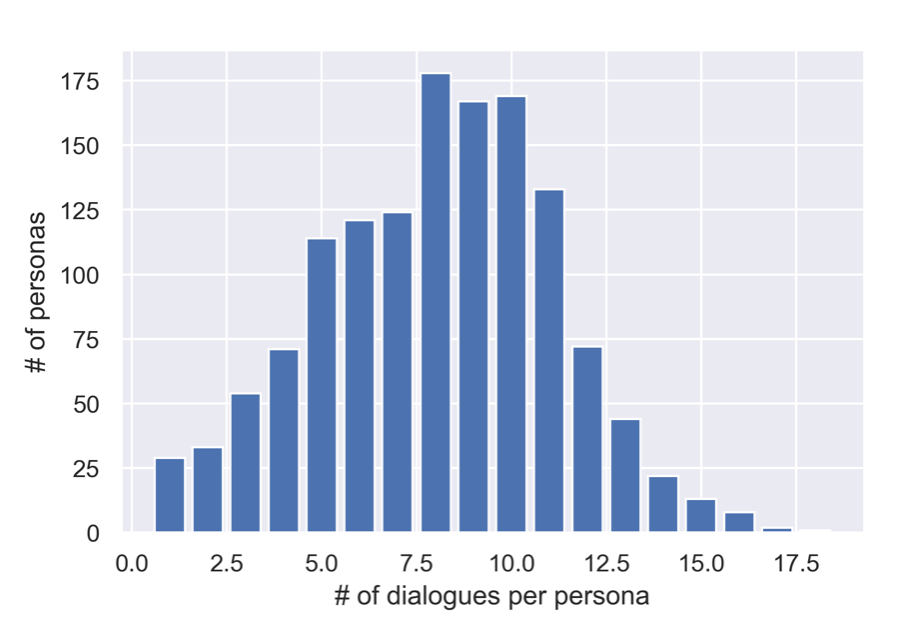}
         \caption{Distribution of the number of dialogues per persona in PERSONA-CHAT}
         %\Description{}
      \label{fig:distribution}
\end{figure*}

\subsection{Distribution of the number of dialogues per persona (Fig~\ref{fig:distribution})}

\subsection{Generated response examples (Table~\ref{tb:examples})}

\begin{table*}
\caption{Generated examples}
\resizebox{1.01\textwidth}{!}{\begin{tabular}{cl}
\hline
\multicolumn{2}{c}{\textbf{Persona}}                                                                                                                                                                                                                                                                                                                                                                                                                                                                                                                                                                                                                                                                                                                                                                                                                \\ \hline
\multicolumn{2}{c}{\begin{tabular}[c]{@{}c@{}}my favorite color is purple\\ i am a 1st grade teacher\\ i have a 3 year old\\ my father is a preacher\\ i go to church on sunday\end{tabular}}                                                                                                                                                                                                                                                                                                                                                                                                                                                                                                                                                                                                                                                       \\ \hline
\multicolumn{2}{c}{\textbf{Dialogue history}}                                                                                                                                                                                                                                                                                                                                                                                                                                                                                                                                                                                                                                                                                                                                                                                                       \\ \hline
\multicolumn{2}{c}{\begin{tabular}[c]{@{}c@{}}Speaker 1: well i do not have any kids yet just four cats .\\ Speaker 2: i would love to see a purple cat ! its my favorite color\\ Speaker 1: oh my favorite color is blue . any hobbies ?\end{tabular}}                                                                                                                                                                                                                                                                                                                                                                                                                                                                                                                                                                                             \\ \hline
\multicolumn{2}{c}{\textbf{Target utterance}}                                                                                                                                                                                                                                                                                                                                                                                                                                                                                                                                                                                                                                                                                                                                                                                                       \\ \hline
\multicolumn{2}{c}{Speaker 2: not really . i really love church though , maybe because my dad is a preacher . you ?}                                                                                                                                                                                                                                                                                                                                                                                                                                                                                                                                                                                                                                                                                                                                \\ \hline
\multicolumn{2}{c}{\textbf{Generated Responses}}                                                                                                                                                                                                                                                                                                                                                                                                                                                                                                                                                                                                                                                                                                                                                                                                    \\ \hline
\multicolumn{1}{l|}{\begin{tabular}[c]{@{}l@{}}Persona + Fine-tuning:\\  Fine-tuning: \\ Persona\_id + Fine-tuning: \\ Reptile + Fine-tuning: \\ Rand init + Prefix-tuning:\\ PersonaPKT (base):\\ PersonaPKT (temperature):\\ PersonaPKT (PPReptile):\end{tabular}} & \begin{tabular}[c]{@{}l@{}}i like to go to church on sunday.\\ i like to go to the beach.\\ i like to play video games.\\ i run an tumblr blog dedicated to purple things it is my favorite color.\\ with my dad. he is a preacher. i am a preacher. i am a preacher. i am a preacher...\\ i go to church every sunday at sunday\\ i do not have any hobbies, i am a teacher. how about you?\\ i do not have any hobbies at all. i am a teacher\end{tabular} \\ \hline
\end{tabular}}
\label{tb:examples}
\end{table*}

\end{document}